\documentclass{article}

\usepackage{PRIMEarxiv}
\usepackage[utf8]{inputenc}
\usepackage[T1]{fontenc}
\usepackage{graphicx}
\usepackage{booktabs}
\usepackage{amsmath}
\usepackage{amssymb}
\usepackage{amsfonts}
\usepackage{array}
\usepackage{multirow}
\usepackage{tabularx}
\usepackage{xspace}
\usepackage{microtype}
\usepackage{fancyhdr}
\usepackage[breaklinks,colorlinks,citecolor=blue,linkcolor=blue,urlcolor=blue]{hyperref}
\usepackage[capitalise,nameinlink]{cleveref}

\graphicspath{{figures/}}

\newcommand{\method}{Bio-MemArt\xspace}
\newcommand{\memart}{MemArt\xspace}
\newcommand{\owner}{Owner\xspace}
\newcommand{\nonowner}{Non-owner\xspace}

\begin{document}

\title{Bio-MemArt: Biometric-Aware KV Cache Memory for Multi-User LLM Agents}

\author{
Yanhong Qian, Xuanying He, Qingguo Meng, Shihao Ding, Xingbo Dong, and Zhe Jin\\
Anhui Provincial Key Laboratory of Secure Artificial Intelligence,\\
School of Artificial Intelligence, Anhui University, Hefei 230093, China\\
\texttt{yanhongqian@stu.ahu.edu.cn; w126221228@stu.ahu.edu.cn;}\\
\texttt{mqg1024@163.com; shihaoding@stu.ahu.edu.cn;}\\
\texttt{xingbo.dong@ahu.edu.cn; jinzhe@ahu.edu.cn}
\And
Hanrui Wang and Isao Echizen (Senior Member, IEEE)\\
National Institute of Informatics, Tokyo 101-8430, Japan\\
\texttt{hanrui\_wang@nii.ac.jp; iechizen@nii.ac.jp}
}

\maketitle

\begin{abstract}
KV cache is evolving from a serving optimization into an external memory substrate for long-term LLM agents.
In a shared multi-user deployment, however, reusable KV blocks introduce a missing access-control question: semantic relevance alone cannot determine whether a memory block is authorized for the current physical user.
We propose Bio-MemArt, a biometric-aware KV-cache memory framework for multi-user LLM agents.
Bio-MemArt attaches a normalized biometric template to each stored KV memory block, filters the shared memory pool with the current user's biometric probe, and then runs the original \memart retrieval and KV reuse pipeline only inside the authorized candidate pool.
This design preserves latent-space retrieval, direct cache reuse, and decoupled position encoding while adding physical-user access control to shared KV memory.
We evaluate Bio-MemArt under \owner and \nonowner query conditions on long-term dialogue QA with face and palmprint benchmarks.
Across face benchmarks, the average owner and non-owner biometric success rates are 95.71\% and 0.86\%; across palmprint benchmarks, they are 97.60\% and 2.00\%.
In the efficiency study, average prefill tokens drop from 18,781.96 under full-context prompting to 28.57 with Bio-MemArt, showing that biometric gating preserves the low-token operating regime of KV-cache memory.
\end{abstract}


\section{Introduction}
\label{sec:intro}
Large language model (LLM) agents are increasingly deployed as persistent assistants that must preserve user-specific state across many turns and sessions.
In this setting, KV cache is no longer only a serving artifact.
Recent work has started to treat reusable KV states as an external memory substrate for agentic systems, because KV blocks preserve model-native hidden states and avoid repeatedly rebuilding long contexts from scratch \cite{memart,keep,agent_below_prompt}.
When historical context is already stored as reusable KV blocks, the system can keep memory in the model-native format instead of reconstructing it from text at every turn.

Shared deployments raise a new problem.
A smart home assistant, public terminal, classroom device, or enterprise agent may serve multiple physical users through one active system.
If their historical KV blocks are stored in a shared memory pool, semantic retrieval alone can still surface another user's private memory.
The core issue is therefore not only which KV block is relevant, but also whether that block is authorized for the current physical requester.

The broader KV-cache literature does not resolve this question.
System work improves serving efficiency through better paging, scheduling, offloading, and storage management \cite{vllm,mooncake,tokencake,dualpath}.
Multi-agent reuse work shares KV states across agents through overlapping prefixes and collective communication \cite{kvflow,kvcomm,tokendance}.
\memart further elevates reusable KV blocks into an explicit long-term memory substrate for LLM agents \cite{memart}.
Yet these methods all optimize how KV blocks are stored, moved, reused, or retrieved; none decides whether a retrieved block should be accessible to the current physical user.
This leaves a deployment gap between memory efficiency and memory authorization in real multi-user systems, where access control must be enforced before semantic KV retrieval can proceed.

Biometric identity provides the missing access signal.
Face and palmprint embeddings can verify who is currently issuing the query more directly than account metadata alone \cite{arcface,ccnet,palm_survey}.
The challenge is to add this identity gate without breaking the benefits that make KV-cache memory attractive in the first place: latent-space retrieval, direct cache reuse, and low prefill cost.

We propose \method, a biometric-aware KV-cache memory framework for multi-user LLM agents, built on the original \memart pipeline \cite{memart}.
Each stored KV memory block keeps the original \memart fields, including the KV block, compressed key, and timestamp, and adds a biometric template vector.
At query time, the current user's biometric probe is compared with the template attached to each KV block.
Only blocks whose cosine similarity exceeds a statistically estimated threshold enter the authorized candidate pool.
The original \memart retrieval then selects top-ranked KV blocks from this authorized candidate pool and reuses them through decoupled position encoding.
Identity filtering therefore happens before semantic KV retrieval, while generation remains unchanged.

This paper makes three contributions.
First, we propose Bio-MemArt, a biometric-aware KV-cache memory framework for multi-user LLM agents, which extends the original \memart memory architecture with physical-user identity cues.
Second, we design an identity-aware retrieval method that attaches biometric templates to KV memory blocks and filters the shared memory pool before native \memart retrieval and KV reuse.
Third, experiments on long-term dialogue QA with face and palmprint benchmarks show that this design preserves strong \owner--\nonowner separation while retaining the low-token advantage of KV-cache memory.

\section{Related Work}
\label{sec:related}

\subsection{KV Cache Systems}
\label{sec:related_kv}

Transformer inference stores key and value tensors for previous tokens so that later decoding steps do not recompute all past states.
Modern inference systems exploit this property through paging, prefix caching, and cache compression to reduce prefill cost and latency \cite{vllm,prefix_cache}.
Recent systems extend this line to agentic and multi-agent workloads, where one deployment may need to keep many partially active agents alive at the same time.
In such settings, the focus is on avoiding fragmentation, keeping GPU memory utilized, and moving KV states without repeated recomputation.

Mooncake pools CPU, DRAM, SSD, and network resources into a KV-cache-centric disaggregated architecture for large-scale serving \cite{mooncake}.
TokenCake studies space contention and time underutilization in multi-agent applications, combining dynamic memory partitioning with event-driven offload and predictive upload \cite{tokencake}.
KVFlow replaces LRU-style eviction with workflow-aware prefix caching guided by an Agent Step Graph and steps-to-execution estimates \cite{kvflow}.
DualPath observes that agentic inference can become storage-I/O bound and introduces dual-path KV loading to use both prefill-side and decode-side bandwidth \cite{dualpath}.
Taken together, these systems establish that KV cache is becoming a first-class systems resource for agent workloads.

However, they still optimize throughput, memory utilization, and data movement.
They do not ask who should be allowed to activate a reusable KV block once that block exists in a shared memory pool.
Our setting starts exactly from this missing question: when reusable KV states persist across users, efficient cache management alone is not enough, because the system also needs an identity-aware access rule before reuse happens.

\subsection{KV Reuse and Memory}
\label{sec:related_reuse}

Another line of work asks how KV states can be reused across contexts or across agents instead of being managed request by request.
KVCOMM aligns overlapping contexts under different prefixes through an anchor pool that estimates cache offsets online, enabling cross-context KV reuse in multi-agent workflows \cite{kvcomm}.
TokenDance exploits the All-Gather communication pattern in synchronized multi-agent rounds and performs collective KV sharing with diff-aware storage \cite{tokendance}.
For scenario-specific deployments, KEEP redesigns KV memory management for embodied planning, where memory updates are frequent and structured \cite{keep}, while Agent Memory Below the Prompt persists quantized KV caches to disk for edge multi-agent inference \cite{agent_below_prompt}.
These works show that KV states can be treated as reusable assets across steps, contexts, or agents.

\memart takes the next step from serving optimization to agent memory itself by treating reusable KV blocks as the external memory substrate of long-term agents \cite{memart}.
This shift is especially relevant to our paper, because once KV cache becomes the memory substrate of a long-term assistant, reuse also determines what personal history can be surfaced in later interactions.
Our work builds directly on this paradigm, but changes the central question from how to reuse more KV blocks to which blocks should be legally accessible in a shared multi-user memory pool.
In other words, existing KV-memory work explains how to preserve and retrieve useful states, whereas our contribution adds an authorization layer that decides whether those states should be exposed to the current requester at all.

\subsection{Biometric Access Control}
\label{sec:related_bio}

Personalized memory becomes risky when multiple users share one device or active session.
Logical identifiers such as account names and conversation IDs can be shared, stale, or spoofed, while the active physical user may change across turns.
Biometrics offer complementary identity evidence because face and palmprint traits are tied to the person making the request.
This makes them attractive when the system must distinguish the current physical user rather than only the nominal account holder.

Face recognition methods such as ArcFace learn discriminative embeddings for verification \cite{arcface}, while palmprint recognition captures line, texture, and structural cues that are useful for contact or contactless matching \cite{ccnet,palm_survey}.
Most biometric work, however, stops at recognition or verification itself.
It does not address how biometric evidence should interact with reusable internal memory states inside an LLM agent.
\method applies biometric verification to KV-cache memory, where the main challenge is not recognition alone, but preserving the retrieval and reuse properties of shared KV blocks after identity gating is introduced.
Our paper differs from both classic biometric verification and prior KV-memory systems by using biometrics as the front-end filter that controls access to reusable KV memory.

\section{Method}
\label{sec:method}

\subsection{Problem Setting}
\label{sec:problem}

We consider a shared LLM agent with a memory pool containing records from multiple users.
For each user $u$, the agent stores historical interactions as memory blocks.
At query time, the system receives a natural-language query $x$ and a biometric probe embedding $b_q$ from the current physical user, and must retrieve useful memory only when that memory belongs to the same user as $b_q$.
We evaluate two conditions: in \owner, $b_q$ matches the biometric template associated with the target memory; in \nonowner, it comes from a different user and should not unlock that memory.
With this setting in place, we next define how biometric identity is attached to KV memory.

\subsection{Biometric-Aware KV Memory}
\label{sec:bio_aware_kv}

\method preserves the original \memart memory representation and augments each KV block with a biometric template \cite{memart}.
For the $i$-th memory block, we store
\begin{equation}
  m_i = \{K_i, V_i, c_i, t_i, a_i, b_i\},
  \label{eq:memory_record}
\end{equation}
where $K_i$ and $V_i$ are the key and value tensors of the historical context block, $c_i$ is the compressed representative key used by \memart for efficient indexing, $t_i$ is a timestamp, $a_i$ denotes auxiliary metadata, and $b_i$ is the biometric template embedding of the user who owns the block.
The biometric template can be produced by a face encoder or a palmprint encoder.
All biometric embeddings are L2-normalized, so matching is performed with cosine similarity:
\begin{equation}
  s(b_q,b_i) = b_q^\top b_i.
  \label{eq:cosine}
\end{equation}
All KV blocks from the same user share the same stored biometric template in our evaluation protocol.
This keeps the identity field lightweight and avoids modifying the KV tensor structure.

\subsection{Threshold Estimation}
\label{sec:threshold}

\method estimates a benchmark-specific biometric threshold from verification data rather than choosing one manually.
For each biometric benchmark, we compute cosine similarities for matched pairs and mismatched pairs.
Let $P^+$ denote the positive similarity distribution and $P^-$ denote the negative similarity distribution.
The threshold $\tau$ is selected at the crossing region between the two distributions:
\begin{equation}
  \tau = \arg\min_z \left| \hat{p}^{+}(z) - \hat{p}^{-}(z) \right|,
  \label{eq:threshold}
\end{equation}
where $\hat{p}^{+}$ and $\hat{p}^{-}$ are estimated density curves.
The resulting $\tau$ is fixed for all downstream memory retrieval trials on the same biometric benchmark.
This avoids tuning the threshold on the dialogue QA task and separates biometric verification from memory evaluation.

\subsection{Identity-First KV Retrieval}
\label{sec:retrieval}

\begin{figure}[tb]
  \centering
  \includegraphics[width=\linewidth]{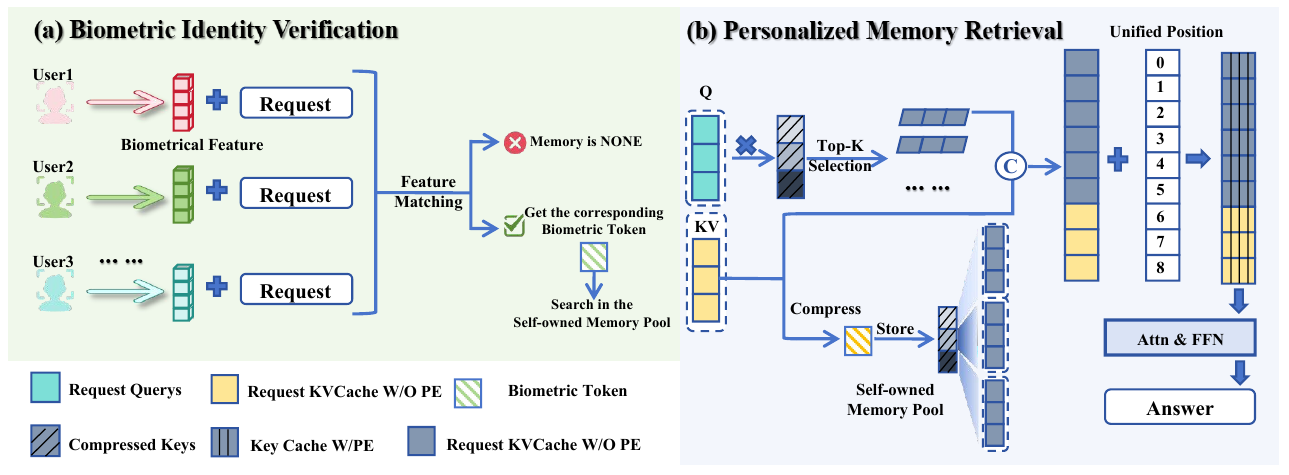}
  \caption{Overall pipeline of \method.
  (a) Biometric identity verification matches the current user's biometric feature against enrolled identities and activates only the corresponding owner memory pool; unmatched requests receive no authorized memory.
  (b) Personalized KV-memory retrieval is then performed inside the authorized candidate pool: the request is encoded into query KV states, compressed keys are used for top-$K$ block selection, selected memories are merged under a unified position layout, and the decoder reuses the retrieved KV cache to generate the final answer.}
  \label{fig:framework}
\end{figure}

\Cref{fig:framework} shows the overall \method pipeline.
Retrieval proceeds in two stages.
First, biometric filtering constructs a query-specific authorized candidate pool:
\begin{equation}
  \mathcal{M}_{bio}(b_q) =
  \{m_i \in \mathcal{M} \mid s(b_q,b_i) \geq \tau \}.
  \label{eq:bio_pool}
\end{equation}
If the probe belongs to the memory owner, the owner's blocks remain available.
If the probe belongs to a different user, the target blocks are removed, preventing their use during generation.

Second, \method{} applies the native \memart{} retrieval function within $\mathcal{M}_{bio}$.
Inside this filtered pool, we keep the original \memart design unchanged, including compressed-key indexing, multi-token aggregation retrieval, and decoupled position encoding for safe KV reuse \cite{memart}.
Biometric filtering only restricts which blocks are eligible to enter the candidate set; it does not redefine the internal retrieval score or the original KV-reuse equations of \memart.
No model weights, attention layers, or decoding rules are changed.

\subsection{Why KV Cache as Memory?}
\label{sec:why_kv}

\method adopts KV cache memory because the reusable unit is already the model's hidden state rather than a re-materialized prompt segment \cite{vllm,prefix_cache,memart,keep,agent_below_prompt}.
This lets the system enforce authorization before large-scale cache reuse happens.
We therefore evaluate not only owner/non-owner separation, but also whether biometric gating preserves the efficiency benefits of KV memory.
This choice also keeps the comparison aligned with the actual deployment question in this paper.
If authorization were added only after converting memory back into text, the system would no longer test whether access control can coexist with native KV reuse.
Using KV memory directly allows us to evaluate both properties at once: whether unauthorized memory can be blocked, and whether the model still benefits from low-prefill hidden-state reuse when access is granted.

\section{Experiments}
\label{sec:experiments}

\subsection{Datasets}
\label{sec:datasets}

\begin{table}[t]
\centering
\small
\caption{Face recognition evaluation datasets.}
\label{tab:face-datasets}
\begin{tabular*}{\linewidth}{@{\extracolsep{\fill}}lccc@{}}
\toprule
Dataset & Type / Group & \#Subjects & \#Images \\
\midrule
AgeDB-30~\cite{moschoglou2017agedb} & Age & 568 & 16,488 \\
CALFW~\cite{zheng2017cross} & Age & 5,749 & 12,174 \\
CFP-FF~\cite{sengupta2016frontal} & Frontal & 500 & 7,000 \\
CFP-FP~\cite{sengupta2016frontal} & Pose & 500 & 7,000 \\
CPLFW~\cite{zheng2018cross} & Pose & 5,749 & 11,652 \\
LFW~\cite{huang2008labeled} & Frontal & 5,749 & 13,233 \\
VGG2-FP~\cite{cao2018vggface2} & Pose & 9,131 & $\sim$3.31M \\
\bottomrule
\end{tabular*}
\end{table}

\begin{table}[t]
\centering
\small
\caption{Palmprint recognition evaluation datasets.}
\label{tab:palm-datasets}
\begin{tabular*}{\linewidth}{@{\extracolsep{\fill}}lccc@{}}
\toprule
Dataset & Type / Group & \#Subjects & \#Images \\
\midrule
\textit{CasiaM\_460}~\cite{sun2005ordinal} & 460nm & 200 & 1,200 \\
\textit{CasiaM\_700}~\cite{sun2005ordinal} & 700nm & 200 & 1,200 \\
\textit{CasiaM\_850}~\cite{sun2005ordinal} & 850nm & 200 & 1,200 \\
IITD~\cite{kumar2008incorporating} & Contactless & 460 & 2,300 \\
\textit{MS\_Blue}~\cite{zhang2009online} & Blue & 500 & 6,000 \\
\textit{MS\_Green}~\cite{zhang2009online} & Green & 500 & 6,000 \\
\textit{MS\_NIR}~\cite{zhang2009online} & NIR & 500 & 6,000 \\
\textit{MS\_Red}~\cite{zhang2009online} & Red & 500 & 6,000 \\
PolyU~\cite{zhang2003online} & Contact & 378 & 7,560 \\
Tongji~\cite{zhang2017towards} & Contactless & 600 & 12,000 \\
\bottomrule
\end{tabular*}
\end{table}

\textbf{Conversational memory benchmarks.}
We use LoCoMo as the primary long-term dialogue QA benchmark.
LoCoMo contains long conversations spanning many sessions and multiple question types that test different forms of memory use \cite{locomo}.
We focus on memory-intensive question types: Multi-Hop, Temporal, and Single-Hop.
These categories directly test whether the agent can retrieve personal facts, combine information across memory entries, and reason about event order \cite{locomo}.
They therefore provide a controlled downstream setting for measuring whether biometric gating changes memory access without changing the QA task itself.

\textbf{Face benchmarks.}
Face-based identity evaluation uses seven public verification benchmarks covering age, frontal-view, and pose variation.
AgeDB-30 and CALFW emphasize age changes; CFP-FF and LFW are largely frontal; CFP-FP, CPLFW, and VGG2-FP are more pose-challenging.
A pre-trained face encoder extracts L2-normalized embeddings, and each benchmark has its own operating threshold.
Table~\ref{tab:face-datasets} summarizes the face datasets, their variation type, and their scale.
Together, these benchmarks let us test whether owner/non-owner separation remains stable under different age and pose conditions.

\textbf{Palmprint benchmarks.}
Palmprint evaluation uses ten protocols from five public datasets.
CasiaM and MS provide wavelength-specific settings; IITD and Tongji are contactless; PolyU is contact-based.
Table~\ref{tab:palm-datasets} summarizes the palmprint protocols and their acquisition settings.
Together with face recognition, these protocols test whether the biometric gate generalizes across complementary physiological traits rather than overfitting to one biometric modality.

\subsection{Evaluation Protocol}
\label{sec:protocol}

The core evaluation principle is to keep the downstream QA task fixed while changing only the biometric operating condition.
For each benchmark, we build the shared memory pool from the same set of LoCoMo users and ask the same questions under both \owner and \nonowner settings.
As a result, changes in F1 or BLEU do not come from different dialogue content; they come from differences in which KV memory blocks remain accessible after biometric filtering.

This protocol explains why some datasets can produce similar QA rows.
When two biometric benchmarks induce the same effective owner acceptance and non-owner rejection pattern on the same LoCoMo split, they can produce similar downstream answers.
Each reported owner/non-owner comparison uses the same stored dialogue history, target user, and question set; only the biometric probe changes.

\subsection{Baselines}
\label{sec:baselines}

We compare against two reference systems.
\textbf{Full-context inference} serves as the upper-cost reference.
\textbf{Native KV memory} uses \memart without biometric filtering and is the primary empirical baseline for downstream QA and runtime comparisons \cite{memart}.
For the same shared memory pool, \owner and \nonowner trials use identical dialogue questions and storage contents; only the biometric probe changes.
This comparison isolates three questions: whether biometric gating prevents unauthorized retrieval, whether it preserves native KV-memory answer quality, and whether it retains the efficiency advantage of KV reuse over replaying full context.

\subsection{Evaluation Metrics}
\label{sec:metrics}

We evaluate three aspects.
\textbf{Memory accuracy} is measured using unigram F1 and BLEU-1.
\textbf{Identity isolation} is measured by biometric success rate and by the \owner--\nonowner gap in downstream QA:
\begin{equation}
  \Delta_{\mathrm{F1}} = \mathrm{F1}_{\owner} - \mathrm{F1}_{\nonowner},
  \quad
  \Delta_{\mathrm{BLEU}} = \mathrm{BLEU}_{\owner} - \mathrm{BLEU}_{\nonowner}.
  \label{eq:gap}
\end{equation}
A larger positive gap indicates that matched users can use personal memory while mismatched users cannot.
\textbf{Efficiency} is measured by prefill tokens, retrieval time, biometric matching overhead, response time, and generated tokens, which are standard deployment-facing indicators for long-context and KV-cache systems \cite{vllm,prefix_cache,memart}.
The central efficiency claim is that \method should remain close to native \memart while requiring far fewer prefill tokens than full-context prompting.

\subsection{Implementation Details}
\label{sec:implementation}

All experiments adopt Qwen2.5-3B-Instruct unless specified otherwise. The face and palmprint pipelines rely on pre-extracted biometric embeddings. For every benchmark, we calibrate the threshold $\tau$ using positive and negative verification pairs and fix this value for all LoCoMo trials.

Our multi-user environment restricts each shared memory pool to at most five users. We perform one-to-one assignment between five biometric identities from the target benchmark and five LoCoMo users, where each user has one biometric template and one independent KV-memory partition. Under the \owner setting, the probe matches the target user’s registered identity; under \nonowner, the probe belongs to another enrolled identity and should be denied memory access. The downstream QA dataset is unified across all benchmarks, so performance gaps only arise from biometric decision thresholds and corresponding accept/reject behaviors.

For the lightweight efficiency comparison in \cref{sec:efficiency_results}, we sample five non-adversarial LoCoMo questions from ten distinct conversations to build 50 QA instances in total, with the generation length capped at 20 tokens. Unless otherwise noted, all storage, compression, retrieval and inference configurations of \memart are kept identical to the original setup in \cite{memart}. The biometric pipelines only supply authorization embeddings, while the rest of the memory pipeline is shared.
This keeps the evaluation focused on what changes when physical-user identity is inserted into KV-memory retrieval without rewriting the long-term memory mechanism.

\section{Results and Analysis}
\label{sec:results}

We report paired owner/non-owner evaluation for face and palmprint benchmarks separately because the two modalities use different verification thresholds.
Because all benchmarks share the same downstream LoCoMo protocol, QA tables should be read together with the dataset-specific authentication summary.

\subsection{Reading Benchmark Differences}
\label{sec:reading_results}

The most important point in this paper is that benchmark differences do not always appear first as large changes in downstream QA numbers.
Because every benchmark is evaluated on the same LoCoMo conversations and the same question subsets, two datasets can produce nearly identical answer scores whenever they induce the same owner/non-owner authorization pattern.
In that case, the biometric benchmark is still affecting the system, but its effect is expressed first through owner acceptance and non-owner rejection rather than through a different language task.

This is why the QA tables and the authentication summary must be interpreted together.
When owner acceptance falls, the authorized user loses access to some relevant memory blocks and the owner QA score drops.
When non-owner acceptance rises, the unauthorized user gains access to memory that should have been blocked and the owner/non-owner gap narrows.
This joint reading explains why some datasets have similar QA rows while still reflecting different biometric difficulty.

\begin{figure}[tb]
  \centering
  \includegraphics[width=\linewidth]{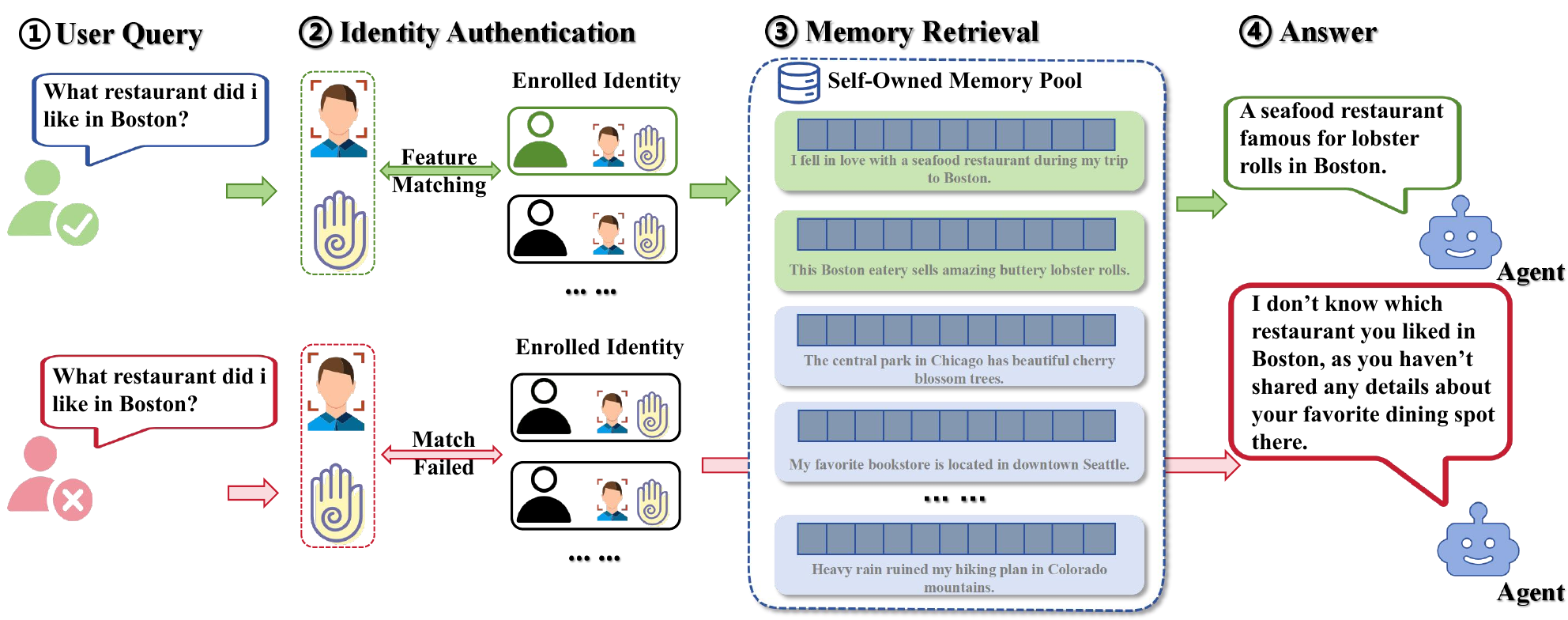}
  \caption{Visualization of the response process under owner and non-owner queries.
  Given the same question, the owner branch passes identity authentication, retrieves only owner memories, and returns the correct personalized answer about the Boston restaurant.
  The non-owner branch fails identity matching, cannot access that user's private memory pool, and returns a safe non-committal response instead of leaking personalized content.}
  \label{fig:case-visualization}
\end{figure}

\subsection{Main Results on Owner versus Non-owner}
\label{sec:owner_results}

The main tables report each dataset with paired \textit{Owner} and \textit{Non-owner} rows.
All entries are F1 / BLEU-1 percentages on the shared LoCoMo Multi-Hop, Temporal, and Single-Hop subsets.
The \textit{Average} column uses the exported \texttt{ALL} split, and \textit{Avg. Gap} denotes the corresponding Owner minus Non-owner difference.
Because all benchmarks use the same downstream QA set, dataset-specific differences are explained by the authentication summary later in this section.

\begin{table}[tb]
  \centering
  \caption{Downstream LoCoMo QA results under face-based biometric-aware KV retrieval.}
  \label{tab:face-main-results}
  \scriptsize
  \setlength{\tabcolsep}{3pt}
  \renewcommand{\arraystretch}{1.03}
  \resizebox{\linewidth}{!}{%
  \begin{tabular}{llccccc}
    \toprule
    Dataset & Condition & Multi-Hop & Temporal & Single-Hop & Average & Avg. Gap \\
    \midrule
    \multirow{2}{*}{AgeDB-30} & Owner & 17.05 / 15.55 & 20.55 / 17.14 & 26.00 / 16.01 & 22.86 / 17.12 & \multirow{2}{*}{16.66 / 12.28} \\
     & Non-owner & 5.55 / 4.55 & 5.37 / 4.63 & 7.47 / 5.43 & 6.20 / 4.84 &  \\
    \midrule
    \multirow{2}{*}{CALFW} & Owner & 14.68 / 13.57 & 20.56 / 17.28 & 24.44 / 14.82 & 21.90 / 16.40 & \multirow{2}{*}{16.59 / 12.24} \\
     & Non-owner & 5.33 / 4.33 & 4.54 / 3.96 & 6.21 / 4.52 & 5.31 / 4.16 &  \\
    \midrule
    \multirow{2}{*}{CFP-FF} & Owner & 17.05 / 15.55 & 21.99 / 18.46 & 26.12 / 16.05 & 23.54 / 17.70 & \multirow{2}{*}{17.34 / 12.86} \\
     & Non-owner & 5.55 / 4.55 & 5.37 / 4.63 & 7.47 / 5.43 & 6.20 / 4.84 &  \\
    \midrule
    \multirow{2}{*}{CFP-FP} & Owner & 17.05 / 15.55 & 21.99 / 18.46 & 26.12 / 16.05 & 23.54 / 17.70 & \multirow{2}{*}{17.34 / 12.86} \\
     & Non-owner & 5.55 / 4.55 & 5.37 / 4.63 & 7.47 / 5.43 & 6.20 / 4.84 &  \\
    \midrule
    \multirow{2}{*}{CPLFW} & Owner & 13.98 / 12.32 & 19.04 / 16.15 & 21.09 / 12.78 & 19.81 / 14.92 & \multirow{2}{*}{14.50 / 10.76} \\
     & Non-owner & 5.33 / 4.33 & 4.54 / 3.96 & 6.21 / 4.52 & 5.31 / 4.16 &  \\
    \midrule
    \multirow{2}{*}{LFW} & Owner & 17.05 / 15.55 & 21.99 / 18.46 & 26.12 / 16.05 & 23.54 / 17.70 & \multirow{2}{*}{18.23 / 13.54} \\
     & Non-owner & 5.33 / 4.33 & 4.54 / 3.96 & 6.21 / 4.52 & 5.31 / 4.16 &  \\
    \midrule
    \multirow{2}{*}{VGG2-FP} & Owner & 15.36 / 13.83 & 21.94 / 18.44 & 23.88 / 14.79 & 22.37 / 16.93 & \multirow{2}{*}{17.06 / 12.77} \\
     & Non-owner & 5.33 / 4.33 & 4.54 / 3.96 & 6.21 / 4.52 & 5.31 / 4.16 &  \\
    \bottomrule
  \end{tabular}
  }
\end{table}

\begin{table}[tb]
  \centering
  \caption{Downstream LoCoMo QA results under palmprint-based biometric-aware KV retrieval.}
  \label{tab:palm-main-results}
  \scriptsize
  \setlength{\tabcolsep}{3pt}
  \renewcommand{\arraystretch}{1.03}
  \resizebox{\linewidth}{!}{%
  \begin{tabular}{llccccc}
    \toprule
    Dataset & Condition & Multi-Hop & Temporal & Single-Hop & Average & Avg. Gap \\
    \midrule
    \multirow{2}{*}{CasiaM\_460} & Owner & 16.40 / 14.91 & 21.55 / 18.11 & 25.51 / 15.78 & 23.01 / 17.35 & \multirow{2}{*}{15.87 / 11.74} \\
     & Non-owner & 6.38 / 5.41 & 6.86 / 5.88 & 7.38 / 5.17 & 7.14 / 5.61 &  \\
    \midrule
    \multirow{2}{*}{CasiaM\_700} & Owner & 15.76 / 14.26 & 21.54 / 18.17 & 24.46 / 15.27 & 22.29 / 16.87 & \multirow{2}{*}{15.92 / 11.95} \\
     & Non-owner & 6.52 / 5.28 & 5.29 / 4.51 & 7.13 / 4.97 & 6.37 / 4.92 &  \\
    \midrule
    \multirow{2}{*}{CasiaM\_850} & Owner & 16.40 / 14.91 & 21.11 / 17.82 & 25.27 / 15.72 & 22.52 / 17.00 & \multirow{2}{*}{16.58 / 12.38} \\
     & Non-owner & 5.33 / 4.33 & 5.42 / 4.60 & 6.34 / 4.49 & 5.94 / 4.62 &  \\
    \midrule
    \multirow{2}{*}{IITD} & Owner & 16.94 / 15.49 & 21.54 / 18.17 & 25.40 / 15.76 & 22.85 / 17.25 & \multirow{2}{*}{16.65 / 12.41} \\
     & Non-owner & 5.55 / 4.55 & 5.37 / 4.63 & 7.47 / 5.43 & 6.20 / 4.84 &  \\
    \midrule
    \multirow{2}{*}{MS\_Blue} & Owner & 17.05 / 15.55 & 21.99 / 18.46 & 26.12 / 16.05 & 23.54 / 17.70 & \multirow{2}{*}{18.23 / 13.54} \\
     & Non-owner & 5.33 / 4.33 & 4.54 / 3.96 & 6.21 / 4.52 & 5.31 / 4.16 &  \\
    \midrule
    \multirow{2}{*}{MS\_Green} & Owner & 17.05 / 15.55 & 21.99 / 18.46 & 26.12 / 16.05 & 23.54 / 17.70 & \multirow{2}{*}{18.23 / 13.54} \\
     & Non-owner & 5.33 / 4.33 & 4.54 / 3.96 & 6.21 / 4.52 & 5.31 / 4.16 &  \\
    \midrule
    \multirow{2}{*}{MS\_NIR} & Owner & 16.40 / 14.91 & 21.99 / 18.46 & 25.41 / 15.69 & 23.16 / 17.46 & \multirow{2}{*}{17.37 / 12.95} \\
     & Non-owner & 5.33 / 4.33 & 4.99 / 4.25 & 6.44 / 4.58 & 5.79 / 4.51 &  \\
    \midrule
    \multirow{2}{*}{MS\_Red} & Owner & 16.40 / 14.91 & 21.99 / 18.46 & 25.41 / 15.69 & 23.16 / 17.46 & \multirow{2}{*}{17.85 / 13.30} \\
     & Non-owner & 5.33 / 4.33 & 4.54 / 3.96 & 6.21 / 4.52 & 5.31 / 4.16 &  \\
    \midrule
    \multirow{2}{*}{PolyU} & Owner & 17.05 / 15.55 & 21.99 / 18.46 & 26.12 / 16.05 & 23.54 / 17.70 & \multirow{2}{*}{18.23 / 13.54} \\
     & Non-owner & 5.33 / 4.33 & 4.54 / 3.96 & 6.21 / 4.52 & 5.31 / 4.16 &  \\
    \midrule
    \multirow{2}{*}{Tongji} & Owner & 17.05 / 15.55 & 21.99 / 18.46 & 26.12 / 16.05 & 23.54 / 17.70 & \multirow{2}{*}{18.23 / 13.54} \\
     & Non-owner & 5.33 / 4.33 & 4.54 / 3.96 & 6.21 / 4.52 & 5.31 / 4.16 &  \\
    \bottomrule
  \end{tabular}
  }
\end{table}

Matched biometric probes consistently preserve better memory-grounded QA than mismatched probes across both modalities.
Several datasets share nearly identical QA rows because they lead to the same accept/reject pattern on the same LoCoMo questions.
For example, CFP-FF, CFP-FP, and LFW all maintain near-perfect owner access while suppressing non-owner access, so their owner scores remain high and their non-owner scores remain low.
By contrast, CPLFW and VGG2-FP lose some owner QA because harder pose variation lowers owner acceptance, and CasiaM\_460/CasiaM\_700 show slightly higher non-owner QA because they admit more false accepts.
Temporal questions remain the most sensitive category in the non-owner setting, which is consistent with identity-gated retrieval suppressing access to user-specific event sequences \cite{memart}.

\subsection{Dataset-Specific Authentication Summary}
\label{sec:threshold_summary}

Because biometric verification quality differs across benchmarks, \method{} uses a benchmark-specific operating threshold before downstream KV retrieval.
Table~\ref{tab:threshold-summary} summarizes the threshold scan together with the owner and non-owner authentication success rates observed in the reported experiments.
These statistics explain why some benchmarks yield stronger downstream isolation than others under the same retrieval pipeline.

\begin{table}[tb]
  \centering
  \caption{Dataset-specific biometric operating points and authentication outcomes used in the reported experiments. `Threshold' is the refined operating point used before KV retrieval, while `Owner' and `Non-owner' report the corresponding authentication success rates in downstream evaluation.}
  \label{tab:threshold-summary}
  \small
  \setlength{\tabcolsep}{3.5pt}
  \begin{tabular}{lrrrr}
    \toprule
    Dataset & Threshold & Delta & Owner (\%) & Non-owner (\%) \\
    \midrule
    AgeDB-30 & 0.2295 & 0.0605 & 98.00 & 2.00 \\
    CALFW & 0.2143 & 0.0979 & 94.00 & 0.00 \\
    CFP-FF & 0.4721 & 0.2052 & 100.00 & 2.00 \\
    CFP-FP & 0.2634 & 0.1021 & 100.00 & 2.00 \\
    CPLFW & 0.2472 & 0.1521 & 86.00 & 0.00 \\
    LFW & 0.5591 & 0.3416 & 100.00 & 0.00 \\
    VGG2-FP & 0.1711 & 0.0593 & 92.00 & 0.00 \\
    CasiaM\_460 & 0.7520 & -0.0081 & 96.00 & 6.00 \\
    CasiaM\_700 & 0.8195 & 0.0026 & 94.00 & 6.00 \\
    CasiaM\_850 & 0.8222 & 0.0102 & 94.00 & 4.00 \\
    IITD & 0.5302 & 0.0342 & 96.00 & 2.00 \\
    MS\_Blue & 0.7402 & -0.0135 & 100.00 & 0.00 \\
    MS\_Green & 0.7445 & -0.0090 & 100.00 & 0.00 \\
    MS\_NIR & 0.8661 & -0.0021 & 98.00 & 2.00 \\
    MS\_Red & 0.8700 & 0.0236 & 98.00 & 0.00 \\
    PolyU & 0.5819 & 0.0172 & 100.00 & 0.00 \\
    Tongji & 0.7105 & 0.0410 & 100.00 & 0.00 \\
    \bottomrule
  \end{tabular}
\end{table}

The authentication summary makes the benchmark-level variation explicit.
For face benchmarks, owner acceptance ranges from $86.00\%$ on CPLFW to $100.00\%$ on CFP-FF, CFP-FP, and LFW, while non-owner acceptance stays between $0.00\%$ and $2.00\%$.
For palmprint benchmarks, most protocols remain strong on both sides, but CasiaM\_460 and CasiaM\_700 admit the highest non-owner acceptance at $6.00\%$, which explains their weaker isolation than MS\_Blue, PolyU, or Tongji.
This is why some datasets differ even when their QA tables look similar: the LoCoMo questions are fixed, so what changes across benchmarks is the biometric operating point and, consequently, which memories can be retrieved.

\subsection{Runtime: Why Use KV Cache Memory?}
\label{sec:efficiency_results}

\method inherits the KV-cache-centric memory substrate from \memart \cite{memart}.
We report a lightweight token-cost comparison between full-context prompting and KV-based memory in \cref{tab:efficiency-compare}.

\begin{table}[tb]
  \centering
  \caption{Efficiency comparison averaged over the same 50 non-adversarial LoCoMo questions with the Qwen2.5-3B-Instruct backbone. Both \memart and \method remain in the low-token regime, while full-context prompting repeatedly recomputes the entire dialogue history.}
  \label{tab:efficiency-compare}
  \small
  \setlength{\tabcolsep}{6pt}
  \begin{tabular}{l@{\hspace{10pt}}r@{\hspace{14pt}}r@{\hspace{14pt}}r}
    \toprule
    Method & Avg Prefill Tokens & Avg Response Time & Median Response Time \\
    \midrule
    Full-context & 18781.96 & 3.6881 & 3.8854 \\
    MemArt & 35.42 & 2.2092 & 2.1679 \\
    Bio-MemArt & 28.57 & 1.8743 & 1.8772 \\
    \bottomrule
  \end{tabular}
\end{table}

\Cref{tab:efficiency-compare} reports all methods as averages over the same 50 questions.
Full-context prompting requires 18,781.96 prefill tokens on average because every query reprocesses the full dialogue history.
By contrast, \memart reduces the average prefill cost to 35.42 tokens, and \method remains in the same low-token range at 28.57 tokens.
The slightly lower token count of \method is consistent with biometric filtering removing non-matching memory blocks before KV retrieval, so the final reused KV set can be marginally smaller than in native \memart.
Response times also remain close, showing that biometric gating does not compromise the core efficiency advantage of KV-cache-centric memory.
The runtime table shows that biometric verification changes which KV blocks can be reused without destroying the low-prefill operating regime of native KV memory.

\Cref{fig:case-visualization} illustrates the final behavioral difference: matched probes recover the personalized restaurant memory, whereas mismatched probes give a generic, non-leaking answer.

\subsection{Modality and Category Analysis}
\label{sec:ablation}

The category-level tables show that the isolation effect appears across all three question types.
Across both modalities, stronger owner acceptance and lower non-owner acceptance translate into larger owner/non-owner QA gaps.
CPLFW is the hardest face benchmark in our evaluation, while CFP-FF, CFP-FP, and LFW preserve the strongest owner access.

For palmprint, CasiaM\_460 and CasiaM\_700 admit more non-owner access than MS\_Blue, PolyU, or Tongji, so their isolation is weaker under the same downstream questions.
Face benchmarks differ mainly in owner retention: stronger pose or age variation reduces the number of correctly admitted owner memories.
Palmprint benchmarks differ more in false acceptance: protocols such as CasiaM\_460 and CasiaM\_700 admit more non-owner access, which narrows the owner/non-owner gap.

Across categories, the weakest non-owner performance appears on questions that depend on personally grounded event sequences, showing that the gate suppresses coherent personalized memory traces rather than only a few isolated facts.
Single-Hop questions rely on direct access to user facts, Multi-Hop questions require combining information across more than one memory item, and Temporal questions depend on ordered event traces.
When non-owner access is denied, all three categories degrade, but the drop is especially visible on questions that require a coherent user-specific memory chain rather than one isolated detail.
That behavior matches the mechanism of Bio-MemArt: once the owner's KV blocks are removed from the authorized candidate pool, the model loses access not only to specific facts, but also to the hidden-state context needed to reconstruct a personalized sequence of events.

\subsection{Practical Interpretation}
\label{sec:practical_interpretation}
Bio-MemArt changes which memory blocks are eligible before native KV retrieval begins. The owner/non-owner gap should therefore be read as an authorization effect on memory eligibility rather than as a general change in language-model behavior.
Strict owner access preserves high-quality authorized KV entries and strong downstream QA, whereas higher non-owner acceptance weakens separation by admitting memory that should have been denied.

Identical QA rows do not imply equivalent biometric benchmarks. Because the LoCoMo task is fixed, what varies across benchmarks is the biometric operating point that determines which hidden-state memories survive the gate. This is why the authentication summary is needed alongside the QA tables.

The efficiency result follows the same logic. Bio-MemArt preserves the original KV reuse pipeline and adds only a selective authorization filter, so it retains low token overhead without falling back to prompt replay.

\section{Conclusion}
\label{sec:conclusion}

We presented \method, a biometric-aware KV-cache memory framework for multi-user LLM agents.
Its identity-aware retrieval method attaches a biometric template to each stored KV block, filters the shared memory pool using the current user's biometric probe, and then applies native \memart retrieval and KV reuse within the authorized candidate pool.
This design addresses the core limitation of shared agent memory: semantically relevant memory is not necessarily authorized memory.
The experiments show three consistent outcomes: authorized users retain substantially better QA performance than unauthorized users, different biometric benchmarks induce different operating points and acceptance rates, and the added identity gate still preserves the low-token operating regime of KV-cache memory.

\bibliographystyle{unsrt}
\bibliography{main}

\end{document}